\pdfoutput=1

\documentclass[11pt]{article}

\usepackage[]{EACL2023}

\usepackage{times}
\usepackage{latexsym}
\usepackage{graphicx}
\usepackage{subfigure}

\usepackage[T1]{fontenc}

\usepackage[utf8]{inputenc}

\usepackage{microtype}

\usepackage{inconsolata}

\title{Automotive Multilingual Fault Diagnosis}

\author{John Pavlopoulos \\ 
  Stockholm University, Sweden \\ 
  \texttt{ioannis@dsv.su.se} \\\And
  Alv Romell\\ Scania, Sweden \\\texttt{alvromell@gmail.com} \\\And
  Jacob Curman\\ Scania, Sweden \\\texttt{curmanjacob@gmail.com} \\\AND
  Olof Steinert\\
  Scania, Sweden \\
  \texttt{olof.steinert@scania.com}\\\And
  Tony Lindgren\\
  Stockholm University, Sweden \\
  Scania, Sweden\\
  \texttt{tony@dsv.su.se}\\\And
  Markus Borg\\
  Lund University, Sweden \\
  \texttt{markus.borg@cs.lth.se}
  \\}

\begin{document}
\maketitle
\begin{abstract}
Automated fault diagnosis can facilitate diagnostics assistance, speedier troubleshooting, and better-organised logistics. Currently, AI-based prognostics and health management in the automotive industry ignore the textual descriptions of the experienced problems or symptoms. With this study, however, we show that a multilingual pre-trained Transformer can effectively classify the textual claims from a large company with vehicle fleets, despite the task's challenging nature due to the 38 languages and 1,357 classes involved. Overall, we report an accuracy of more than 80\% for high-frequency classes and above 60\% for above-low-frequency classes, bringing novel evidence that multilingual classification can benefit automotive troubleshooting management.
\end{abstract}

\section{Introduction}

Fault diagnosis is the task of detecting the fault that caused a problem or unexpected behaviour to a subject. If the subject is a human being and the nature of the problem is medical (e.g., COVID-19), a reasonable diagnostics process comprises the physician who reads or listens to the patient’s symptoms, looks at any radiographs or echocardiograms, studies the medical records, and so on. Then, the physician may conclude regarding the root cause and suggest the right treatment. In the automotive industry, a similar process is followed, because ensuring the functional safety over the product life cycle while limiting maintenance costs has become a major challenge \cite{theissler2021predictive,zhao2021challenges}. Natural Language Processing (NLP) has facilitated medical diagnostics \cite{irving2021using,izquierdo2020clinical} and issue management in engineering contexts, e.g., telecommunications \cite{jonsson2016automated} and banking \cite{aktas2020automated}. In this work, by taking a vehicle fleet as a subject, we show that NLP-assisted troubleshooting management is feasible also in the automotive domain. In line with previous work, we show that it can serve as an additional channel to serve corrective maintenance and health management \cite{safaeipour2021survey,theissler2021predictive,nath2021role,vaish2021machine}. 

Upon a (e.g., mechanical) fault a driver typically communicates with the fleet manager, i.e., the one responsible for the vehicles in the company's fleet throughout each vehicle's life cycle. The driver shares the details of the problem as a text message (email, SMS, voice mail, etc.) and the department advises the driver to move the truck to a dedicated support centre (workshop) nearby. An expert is assigned to diagnose the root cause of the fault and when the diagnosis is complete, the problem can be fixed (e.g., by ordering replacement parts) so that the driver can continue the job routine. The time of the aforementioned process is not short. The driver might be suggested a service-centre that is suboptimal for the fault in question or a technician that doesn’t have the right skill set (e.g., high voltage for EVs), which leads to a longer time before the truck is repaired. However, if the problem at hand was known at an early stage, the company could plan accordingly and find an empty slot in a workshop, order spare parts, prepare invoices, etc. 

We study data shared by Scania and attempt to classify the textual descriptions of the problems, as these were registered through work orders in workshops, regarding the actual root cause for the vehicle malfunction. We formed a dataset of 452,071 texts, written in 38 languages and classified manually to 1,357 classes. We then investigated whether large-scale text classification could assist with faster resolution of faults, hence leading to a better working environment for drivers and mechanics, improved logistics, and better troubleshooting management overall. Although AI-enabled prognostics and health management is a well-studied field \cite{zhao2021challenges}, in this work we show that NLP can open a new path for automotive troubleshooting management in \textit{effectiveness (diagnostics assistance)}, \textit{efficiency (speedier troubleshooting)}, and \textit{management of decision-making trade-offs (better-organised logistics)}.

In specific, the symptoms, which are usually written in natural language (e.g., emails, SMS), before arriving at the company’s front desk could first be passed through a text classifier, trained to detect the fault behind the claim. At this stage, neither the end user nor the company know the problem. The assumption, however, is that a classifier can learn to predict the underlying fault based solely on the textual claim while a system-predicted fault could: (a) assist the mechanics/diagnostics teams toward reaching faster to the root cause of the problem (speedier troubleshooting); (b) reduce the human error, given that the tired or inexperienced expert will be assisted with the system-prediction; (c) allow ordering of parts in a timely manner, by detecting early patterns in the fault reports, hence leading to better organisation of the logistics.

This work presents the first large-scale study that demonstrates the applicability of automated fault report management in the automotive domain. Different from medical fault diagnosis, which is most often based on image input (Appendix~\ref{apx:challenge}), its large-scale multi-class and -linguality nature, combined with the terminology, constitute a specific task and a challenging problem from an NLP perspective. 
In the rest of this article, Section~\ref{sec:related_work} presents the related work and Section~\ref{sec:data} presents the dataset while Section~\ref{sec:methods} provides an empirical analysis. Section~\ref{sec:discussion} discusses the findings under the light of error analysis and is followed by our suggestions for future work.

\section{Related work}\label{sec:related_work}
Fault diagnosis is a well studied problem \cite{safaeipour2021survey}. It is similar to corrective maintenance, defined as the task of repairing a system after a failure occurred \cite{theissler2021predictive}. Fault diagnosis is also related to failure detection and predictive maintenance \cite{Carvalho2019}, and prognostics and health management \cite{nath2021role, biteus2017planning}. Thanks to digitalisation, managing fault reports in information systems has provided organisations new opportunities to increase the level of automation in the related work tasks.

Guided by the ``big data'' mindset, research and practice have successfully used machine learning (ML) to automate fault report management. In large organisations, the inflow of textual fault reports often contains actionable patterns for ML models to learn. The software engineering community was an early adopter of the approach and numerous papers on training classifiers to analyse bug reports have been published. Common applications include duplicate detection, bug prioritisation, and automated team assignment for rapid bug fixing~\cite{borg2014changes}. 

NLP is also used to facilitate business processing, but most often through the development of task-directed dialogue systems (chatbots), e.g., to assist user satisfaction assessment \cite{borsci2022chatbot} or troubleshooting \cite{thorne2017chatbots}. Although machine learning is present in such studies, this is not to assist a diagnostics process but rather tasks such as intend classification \cite{adamopoulou2020overview}. The broader potential of NLP in prognostics and health management, however, is not disregarded, with tasks such keyword detection in maintenance records and prediction of the failure type remedied proposed \cite{fink2020potential}.

Most previous work that attempted to employ NLP to address fault diagnosis, relied on simple bag-of-words models or the TFIDF statistical measure followed by standard techniques available in open-source ML libraries. \citet{jonsson2016automated}, for example, compared most techniques available in WEKA for training classifiers for telecommunications fault reports at Ericsson.\footnote{\url{https://www.cs.waikato.ac.nz/ml/weka/}} \citet{aktas2020automated} presented a similar study for fault reports in the context of İşbank, the largest bank in Turkey. \citet{vaish2021machine} trained various ML classifiers for fault reports in the domain of power systems. Recently, deep learning has also been applied to classify fault reports, including recurrent and convolutional neural networks \cite{zhang2022intelligent} or transfer learning \cite{qian2022deep}. All these approaches, however, are outdated, since BERT \cite{devlin2018bert} set a new state of the art in several NLP tasks. 

This study makes two significant contributions in light of previous work. 
First, we present the first large-scale study that demonstrates the applicability of automated fault report management in the automotive domain, reporting promising results for an industrial case at Scania. 
Second, we show that the state-of-the-art in NLP can handle effectively multi-lingual fault diagnosis, hence unlocking the use of pre-trained masked language Transformer models \cite{devlin2018bert,conneau2019unsupervised} for fault diagnosis in the automotive domain and beyond.

\begin{figure*}[ht]
    \centering
    \includegraphics[width=.8\textwidth]{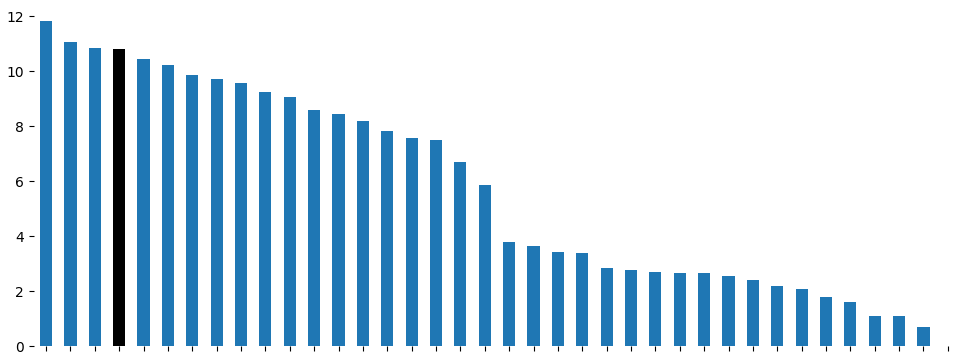}
    \caption{The (log) frequency distribution of the languages in the data. Portuguese, German and English are the most frequent languages of the texts, followed by an undefined language (in black).}
    \label{fig:lang_dist}
\end{figure*}

\section{The multilingual and multiclass dataset}\label{sec:data}
We present the first dataset with textual descriptions of malfunctions in trucks. Approx. 452,071 texts have been extracted from a database containing work orders placed in workshops over the past years. Each text is labelled corresponding to a \textit{main group} and a \textit{class}. The main group refers to what overarching segment the faults belong to (e.g., `engine' or `chassis') and the class reflects the particular sub-part that has been identified as the root cause for the vehicle malfunction in a workshop (e.g., `oil pump' or `yoke'). In all our experiments, we used 52\% of the data for training, 24\% for development and 24\% for testing. The shortest sample contains as little as one word, whereas the longest one consists of over 350 words.

\subsection{Language distribution}
Thirty-eight languages have been identified in the data with the help of the Amazon Translate language detection service.\footnote{\url{https://aws.amazon.com/translate/}} In Figure~\ref{fig:lang_dist}, which shows the most frequent languages, the large imbalance between the languages can be observed. The ten most frequent languages make up 93.3\% of the total samples while the ten least frequently predicted languages make up 0.01\%. 
The ten most frequent languages make up 93.3\% of the total sample and are the following (unordered)\footnote{The order, along with further information, is not revealed to protect sensitive company information.}: English, German, Swedish, Finnish, Norwegian, Danish, French, Dutch, Portuguese, Italian, Czech, Hungarian, Spanish, and Lithuanian. 

When the translation service was not able to identify a language an unknown language (\textsc{unk}) was assigned (highlighted in black in Figures~\ref{fig:lang_dist} and~\ref{fig:classes_per_lan}). 
Through manual inspection, it is possible to identify flaws in language detection and to find translations of cases where a language has been identified while the translation was not accurate. An example is ``110-5002.06 SKARVKOPPLING 6 MM PLAST'', which is Swedish predicted to be Vietnamese; or ``TMI 04 15 02 19'' and ``tpm 48180'', which were incorrectly predicted as Haitian and Indonesian; or the detected language ``Esperanto''. Although we cannot exclude the case that another translation service will handle better such cases, we find that the translation of these texts is hard.

\subsection{Class distribution}
\begin{figure}[ht]
    \centering
    \includegraphics[width=.4\textwidth]{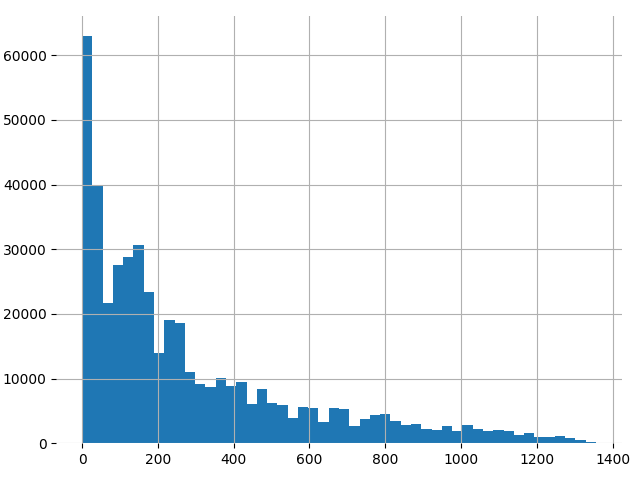}
    \caption{Histogram of the classes.}
    \label{fig:class_dist}
\end{figure}

There are 1,357 unique classes in the available data, each indicating a subpart that has been identified as being the root cause of a problem in a workshop work order. Class examples are `water valve', `yoke', and `oil pump'. The distribution between the classes is heavily imbalanced, because some errors and faults are more likely to appear earlier in the life cycle of a truck than others. The histogram in Fig.~\ref{fig:class_dist} shows the right-skewed distribution.

\begin{figure}[ht]
    \centering
    \includegraphics[width=.45\textwidth]{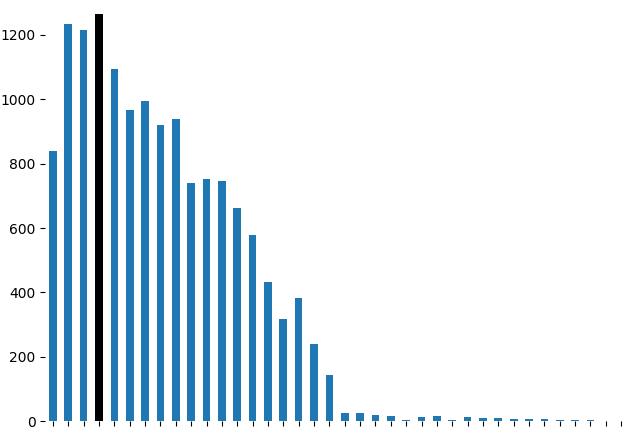}
    \caption{Number of classes per language. The black bar corresponds to an undefined language (\textsc{unk}).}
    \label{fig:classes_per_lan}
\end{figure}

\begin{figure}[ht]
    \centering
    \includegraphics[width=.45\textwidth]{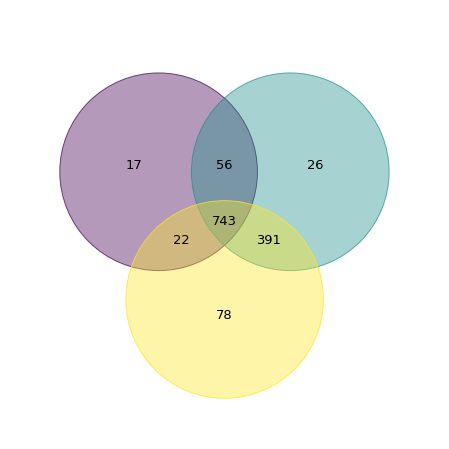}
    \caption{Overlap of represented classes for the three most frequent languages, Portuguese (lowermost), German (right), and English (left).}
    \label{fig:classven}
\end{figure}

The class distribution varies among languages and the number of unique classes that are represented in each individual language differs (Fig.~\ref{fig:classes_per_lan}), with German, English and \textsc{unknown} (i.e., where the translation fails) being the ones with the most classes. The ten most frequent classes make up 14.8\% of total samples, while the one hundred least frequent classes on the other hand make up less than 0.3\%. 
The Venn--diagram in Figure~\ref{fig:classven} shows the class overlap between the three most frequent languages, vis. Portuguese, German and English (see Fig.~\ref{fig:lang_dist}). Although we do not limit the number of classes to those shared by all or certain languages, we note the fact that classes are not represented equally in the languages.

A manual inspection of the data showed that relatively common phrases exist, such as ``customer complaint'', ``driver complaint'', ``attend to'', ``vehicle presenting'', which do not provide any information regarding the actual fault. Inspecting to what extent these phrases are present in the data showed that around 7\% of the training data comprised at least one of these phrases\footnote{These phrases were investigated in the context of translated data, to avoid the cumbersome work of searching in all possible languages.}.

\begin{table}
\label{tab:misclassified}
\label{tab:badsamples}
 \begin{tabular}{ c p{4.6cm} c}
\hline 
\textbf{Group} & \textbf{Original text} & \textbf{Class} \\ 
\hline 
 14 & 110-5002.06 SKARVKOPPLING 6 MM PLAST & 4 \\
 4 & TMI 04 15 02 19 & 38 \\
 8 & Noise from rear axle & 1227 \\
 15 & Luftverlust am Fahrersitz & 223 \\
\end{tabular} 
\caption{Examples taken from the dataset. The predicted language of the topmost example was Norwegian (``110-5002.06 JOINT COUPLING 6 MM PLASTIC'') and of the last it was German (``Air loss at driver's seat'').}
\end{table}

\section{Empirical analysis}\label{sec:methods}
The task of diagnosing a fault from its textual description can be approached as a large-scale text classification problem. Given that the dataset we experimented with comprises texts in multiple languages, the task is a multilingual large-scale text classification (LSTC) problem. Opted text classification methods were multilingual or they operated on the data after these were translated into English (ET). 

\subsection{Benchmarks under the light of translation}
Our study aims to draw the baseline performance for this multiclass task, but also to investigate the benefits of employing translation as a service. Hence, we opted for multilingual methods, as well as for methods operating on texts translated to English. Our main multilingual method was XLM-R (base), a BERT-based model with state-of-the-art performance on multilingual tasks~\cite{conneau2019unsupervised}. Our main method employing ET was a pre-trained DistilBERT model \cite{Sanh2019}, preferred over the original BERT \cite{devlin2018bert} due to its documented performance compared to the original BERT, while being more lightweight and having significantly fewer parameters, thereby reducing training (fine-tuning) time~\cite{sanh2019distilbert,shaheen2020large}. 
We also experimented with a multinomial logistic regression on top of FastText (FTX) embeddings \cite{joulin2016bag} and with a Convolutional Neural Network (CNN) baseline. Although recurrent neural networks work well for the NLP tasks where comprehension of long range semantics is required, CNNs work well where detecting local and position-invariant patterns is required, such as key phrases~\cite{8666928,minaee2021deep}. DistilBERT and CNN were also trained on the English translations. Overall, we experimented with six baselines: FTX-ET, CNN-ET, DistilBERT-ET, DistilBERT, CNN and XLM-R. A majority baseline is also used to draw the task difficulty.

\subsection{Evaluation}
Our task is a multiclass problem with few classes outweighing thousands of others (see Figure\ref{fig:class_dist}). Some of the rare classes occur in only few claims, which are not of great interest for our troubleshooting management use case.\footnote{Scania owns complementing tools to handle rare classes.} By contrast, issues related to fewer yet frequently occurring classes can reveal trends that possibly transfer across countries and can be effectively addressed with early troubleshooting management. Hence, for evaluation purposes, we opt for top-k accuracy, which counts the number of times the true label is among the k classes predicted with the highest probability. To better present and analyse the results of this large-scale multiclass problem, we introduce support and language -based zones, by segmenting the evaluation data based on their class-support and language. We show results with $k=1$, but Appendix~\ref{apx:eval} comprises results with more values, as well as with precision, recall and F1. 

\begin{figure}[ht]
    \centering
    \includegraphics[width=.48\textwidth]{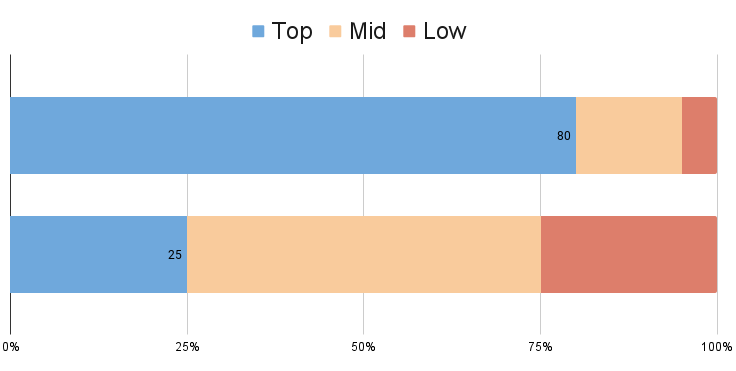}
    \caption{Segmentation based on the number of instances (lowermost), showing the percentage of samples (horizontally) per segment (colored). The same is shown on top when segmenting based on the number of classes.}
    \label{fig:datasegmentation}
\end{figure}

\noindent\textbf{Segmentation based on class support} is performed by clustering the classes based on their size, and then evaluating per cluster. We used the 1st (25\%) and 3rd (75\%) quartile as our two thresholds, in order to yield three class zones, shown with the lowermost bar of Fig.~\ref{fig:datasegmentation}. The low-support zone (in red) comprises 1,076 classes whose total number of instances (classified in one of those classes) does not exceed the first quartile of our data. The top zone (in blue) is similar but using 27 high-support classes. The mid zone comprises the rest. The top and low support zones comprise the same number of texts, in order to set a scene where low-support classes are of similar interest to management as high-support classes. When segmenting based on the number of classes (upper horizontal bar in Fig.~\ref{fig:datasegmentation}), instead of their support (lower), the three zones are heavily imbalanced. 

\noindent\textbf{Language-based evaluation} can be performed by grouping texts of the same language. In this case, we note that while it is possible to calculate the accuracy for a language by dividing the number of true positives in a language by all samples in that language, calculating other metrics, such as precision and recall would require the computation of false positives and false negatives, which is not feasible in the language context. In the multi-class setting each sample which is not classified correctly is both a false negative for the true class and a false positive for another class. This would lead both recall and precision to be identical to accuracy when looking at samples belonging to a single language.

\subsection{Experimental results}

\begin{table}[]
    \centering\small
    \begin{tabular}{lcccc}
          & \sc total & \sc low & \sc mid & \sc top \\\hline
         Majority & 2.46\% & 0.32\% & 1.03\% & 9.80\%\\\hline
         FTX-ET & 19.6\% & 8.5\% & 22.7\% & 24.5\% \\
         CNN-ET & 49.5\% & 22.8\% & 53.0\% & 68.3\% \\
         DistilBERT-ET & 61.4\%& 35.5\%& 62.2\% & 78.8\% \\
         DistilBERT & 61.1\% & 33.0\%& 64.3\% & 82.3\% \\
         CNN & 52.3\% & 22.8\% & 56.3\% & 72.4\% \\
         XLM-R & 61.3\% & 29.8\% & 66.6\% & 82.0\% \\\hline
    \end{tabular}
    \caption{Classification accuracy of models trained on English translations (ET) and on raw data (lowermost). Accuracy on all classes has been computed, as well as on three clusters formed based on class support.}
    \label{tab:results}
\end{table}

Table~\ref{tab:results} presents the accuracy per model. A majority baseline that achieves a very low score shows the task difficulty, with the top-zone being the easiest due to fewer classes being considered. FTX-ET and CNN perform poorly, but with the latter being clearly better. A DistilBERT trained on English translations is the best (61.4\%), despite the fact that approximately 5\% of the instances are missed due to the inability of the translation service to produce a translation. When ignoring these during the evaluation, the accuracy drives up to 61.9\% (more details are shared in Appendix~\ref{apx:unknown}). The multilingual XLM-R follows closely (61.3\%), despite the fact that it does not employ any translation service, operating on texts presented with their original language in which they were written. 
Furthermore, when we train DistilBERT on all (multilingual) data, the performance improves for mid/top classes while it drops in low. This drop can be explained by the fact that there aren't enough data to learn during fine-tuning. By contrast, better-supported classes are better handled without any translation, probably because the model has enough data to learn to trust the terms ignoring the rest.

\subsubsection{Class-based assessment}
The performance per class-support zone is shown in the three rightmost columns of Table~\ref{tab:results}. CNN performs clearly better than FTX-ET but both fall behind the other models. DistilBERT-ET is better than DistilBERT only in the low-support zone. This means that when the class support is low, this monolingual pre-trained masked language model benefits (+2.5) from using English translations as input, instead of the raw data. By contrast, when the class support is higher the translation step is not only redundant, but also harms the results in the mid (-2.1) and in the top zone (-3.5). For frequently occurring classes (top segment, last column of Table~\ref{tab:results}), DistilBERT is the best, followed closely by XLM-R (-0.3) and DistilBERT-ET (-3.5). Using the raw input information, by disregarding the language and the translation, appears to provide a better input signal, which is also shown with the superiority of CNN (72.4\%) over its translation-based counterpart (68.3\%). 

\subsubsection{Language-based assessment}\label{ssec:results_per_lang}

\begin{table}[ht]
\centering
\begin{tabular}{lcc} 
& \textbf{DistilBERT-ET} & \textbf{XLM-R} \\
\hline
1&71.3 & \textbf{72.4}   \\
2&61.3 & \textbf{61.4}   \\
3&\textbf{61.0} & 58.5   \\
4& -- & \textbf{44.3}   \\
5& 48.3 & \textbf{48.5}   \\
6& 52.4 & \textbf{53.2}   \\
7& \textbf{62.8} & 60.5   \\
8& 55.3 & \textbf{54.0}   \\
9& 61.0 & \textbf{61.4}   \\
10& \textbf{57.4} & 57.1   \\
11& \textbf{38.0} & 36.0   \\
12& 62.4 & \textbf{64.8}   \\
13& \textbf{57.8} & 55.5   \\
14& 61.7 & \textbf{62.0}   \\
15& \textbf{55.1} & 50.5   \\
16& 46.2 & \textbf{54.6}   \\
17& \textbf{51.4} & 46.3   \\
18& \textbf{54.5} & 49.7   \\
19& 40.8 & 40.8 \\
\end{tabular}
\caption{Accuracy (\%) per language of the best performing monolingual (DistilBERT-ET) and multilingual (XLM-R) model. Only languages with log frequency above five are considered and DistilBERT is not computed when the language was undefined.}\label{tab:lang_acc}
\end{table}

Table~\ref{tab:lang_acc} presents the Accuracy (\%) per language of the best performing monolingual (DistilBERT-ET) and multilingual (XLM-R) model. Only languages with a log frequency above five are shown, since below that threshold the support significantly drops (Fig.~\ref{fig:lang_dist}). At the same threshold, the number of unique classes is also reduced (Fig.~\ref{fig:classes_per_lan}). 

\noindent\textbf{The five most frequent languages} are all better addressed by the multilingual XLM-R except from English (3rd in row), the language DistilBERT is pre-trained on. XLM-R also performs well (44.3\%) for texts that the translation service fails to provide an English translation (\textsc{unk}), which are texts that DistilBERT is incapable of handling. DistilBERT, however, is better for the majority of the thirteen less frequent languages with 2.8 units on average, ranging from 0.3 added units for the 10th language to 5.1 for the 17th. This unexpected finding shows that a simple lightweight translation-based model, outperforms overall its multilingual counterpart for low-represented languages in this domain. 

\subsection{Oversampling low-represented classes}
\begin{figure}[ht]
    \centering
    \subfigure[]{
    \includegraphics[width=.45\textwidth]{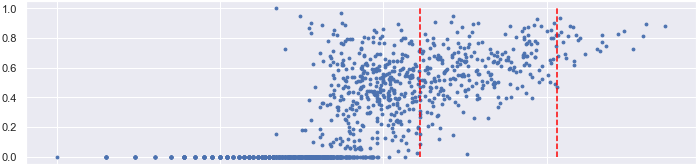}
    }
    \subfigure[]{
        \includegraphics[width=.45\textwidth]{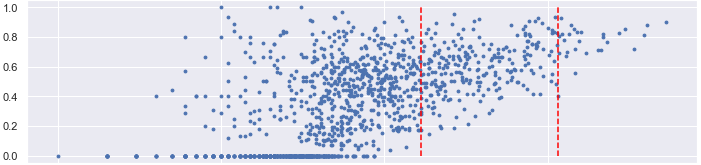}
    }
    \subfigure[]{
        \includegraphics[width=.45\textwidth]{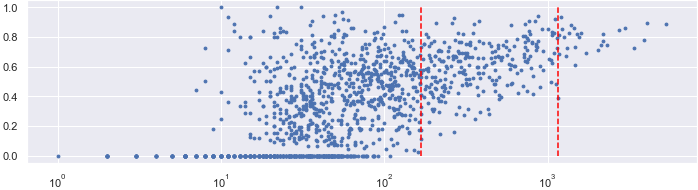}    
    }
    \caption{XLM-R's $\textit{F}_1$-score per class (point) based on class support (horizontally; log10-scaled) when oversampling classes with a support lower than 0 (a), 30 (b) and 50 (c). Red dashed lines separate low (left), mid and top (right) support zones.}
    \label{fig:xlmr_oversampling}
\end{figure}

To study the possibilities of mitigating the effects and learning issues caused by intrinsic and extrinsic factors through the use of simple methods, we oversampled under-represented classes through random sampling with replacement. Classes with fewer instances than a threshold were augmented by duplicating their instances.\footnote{Instances were selected randomly, up to twenty per class. } We varied this threshold from 10 to 50 and found that oversampling classes with less than 30 instances yields the most benefits for the low-segment (+2) without harming any of the other segments and the overall accuracy.\footnote{Instead of oversampling, we also experimented with adding the texts' translations to English, but this approach was overall worse.} Figure~\ref{fig:xlmr_oversampling} shows XLM-R's $\textit{F}_1$-score for each class plotted against its frequency in the training data. When no oversampling is employed (a), in the low-support zone on the left, there is a cut-off point below which classes are incorrectly predicted. When oversampling classes with a support lower than 30 (b) and 50 (c), the same region is more densely populated by classes with high $\textit{F}_1$-values. Indicatively, the total accuracy of XLM-R increased with oversampling from 61.3\% to 62.5\% (threshold equal to 10) to 62.6\% (20) to 62.7\% (30) to 63.2\% (50).

\section{Discussion}\label{sec:discussion}

\begin{figure}[ht]
    \centering
    \includegraphics[width=.45\textwidth]{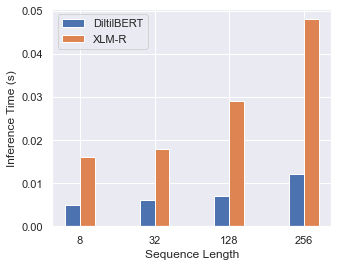}
    \caption{Inference time for different sequence lengths with a batch size of 1 on an Nvidia T4 GPU.}
    \label{fig:inference}
\end{figure}

As was shown in Sect.~\ref{sec:methods}, DistilBERT operating on texts translated to English achieved the best results overall (see Table~\ref{tab:results}). This approach, however, carries the extra cost of translation. The multilingual XLM-R and the monolingual DistilBERT followed closely. Looking at support zones, DistilBERT is the best for high-frequency classes, outperforming the multilingual XLM-R and the translation-based DistilBERT-ET. The vast amount of data for a limited number of classes, make the original languages a better input space compared to English translations, despite the fact that this method is pre-trained on an English corpus. XLM-R is also left behind for the low-zone, outperformed by both DistilBERT models. However, its better performance for the bigger mid zone, makes it the best option overall when translation is not an option. On the other hand, DistilBERT, which is the best in the low and top zones, is also far more lightweight. In specific, the training time of XLM-R is approx. 26 hours while that of DistilBERT is 6. Also, as can be seen in Fig.~\ref{fig:inference}, XLM-R is slower during inference.\footnote{The same finding was verified when we used the CPU. The difference was smaller for an Nvidia Tesla P100 GPU.} 

\begin{figure*} [ht]
    \centering
    \includegraphics[width=\textwidth]{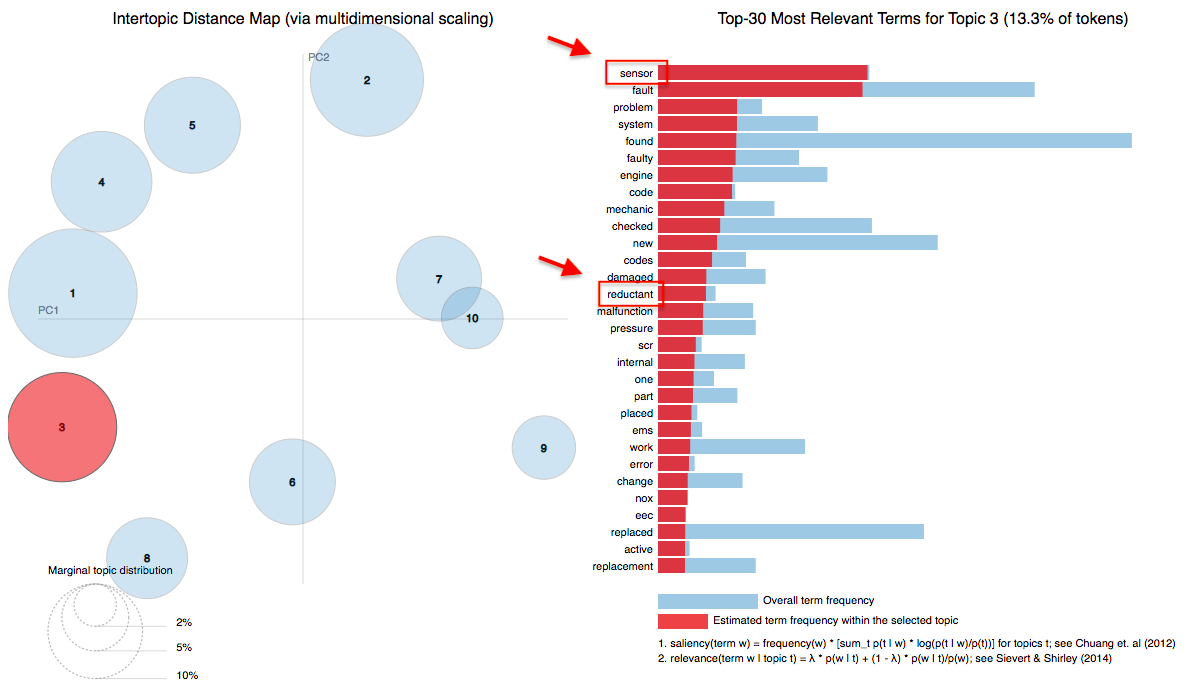}
    \caption{Topic modeling with Latent Dirichlet Allocation on the validation data. The focus in on the third topic, shown in red. Topics are represented by circles whose size reflects the marginal distribution. The term-frequency is shown in blue bars overall and in red for the topic in question.}
    \label{fig:lda}
\end{figure*}
\noindent\textbf{A language-based assessment} (see Table~\ref{tab:lang_acc}) revealed that a multilingual Transformer can work equally well or better than a model operating on texts translated to English. This is important if costs are an important factor while, as already discussed, this comes hand in hand with an improved efficiency. Accuracy ranges from above 70\% for the most frequent class to around 40\% eighteen positions lower in the rank. This score is only five points above the best performing model for the low-zone, which achieved 35.5\%. By oversampling the low-supported classes, we observed that better scores are feasible (Fig.~\ref{fig:xlmr_oversampling}), even with simple mechanisms, such as instance duplicating.

\noindent\textbf{An error analysis} for the most frequent classes revealed that both our best performing models, DistilBERT (monolingual) and XLM-R (multilingual), perform well, correctly classifying most of the instances of the respective class (the confusion matrices can be found in Appendix~\ref{apx:conf}). Although mistakes were detected, these were similar for both models, which confused the following three pairs: (2,12) that stands for `reductant hose' and `NOx sensor', (140, 159) that stands for `sealing ring' and `seal', and (5,16) that stands for `yoke' and `support'. A linguistic analysis revealed that the classes of the two former pairs concerned the same mechanical malfunction, which means that the prediction could have been even higher if we grouped classes together.
In specific, a `sealing ring' is a circular seal, which can be considered a more specific concept (hyponym) of `seal'. A `sealing ring' that is classified as a `seal' is now considered as a mistake. Also, a `yoke' is a yoke-shaped plate or such, on which other components can be attached. It can be considered as a subtype of `support', which is defined as a device that supports and helps hold a unit in a certain position. Finally, a `NOx sensor' and a `reductant hose' are two components of the same system, the exhaust gas after-treatment that lies below fuel and exhaust system. According to a domain expert, the symptoms of a faulty `NOx sensor' and a `faulty reductant hose' are similar, though not identical. An analysis with topic modeling \cite{blei2003latent}, however, revealed that the terms `sensor' and `reductant' co-exist in the same topic (Fig.~\ref{fig:lda}).\footnote{We used: \url{https://pyldavis.readthedocs.io} and Gensim's LDA implementation.} 

\section{Conclusion}
We presented the first large-scale study that demonstrates the applicability of automated fault report management in the automotive domain.
Empirical findings, using data from vehicle fleets, revealed that Transformer-based models can adequately address this large-scale multilingual multiclass text classification task, opening the way for the application of the same workflow on similar domains. Our findings show that translation-based data assist low-represented classes while, when translation is not an option, using a model pre-trained on multilingual data or fine-tuning a model pre-trained on English data can perform equally well, or even better for high-frequency classes. Also, the relatively low performance of low-frequency classes can be improved with oversampling, a direction we plan to investigate further in future work, along with hierarchical classification and the exploitation of more (labelled and unlabelled) data.

\newpage
\section*{Limitations}
The identified limitations of this work, described below, are all currently under investigation.
\begin{itemize}
    \item We ignored ``time locality'', but some errors and faults are more likely to appear earlier in the life cycle of a truck than others. More generally, certain faults might cluster in time, which means that their respective claims will not be independent. For example, if component A brakes in Truck X, perhaps component B is more likely to break next.
    \item We only considered the text of the problem or the symptom, in this study, ignoring any metadata. We note, however, that predictive power may exist in the metadata in the issue management system. The location, the vehicle model, the time of the year, etc. can assist the classification and improve the performance.
    
\end{itemize}
\section*{Ethical considerations}
\begin{itemize}
    \item The data may provide sensitive information of employees. Hence, we don't release the dataset in raw format, but we will release the contextual text embeddings along with class labels, to allow reproducibility and further benchmarks in future work.
    \item The classifiers could in principle be used to assist the compilation of false claims. The class labels, however, are encoded and any released models classify indices and not class names.
\end{itemize}

\bibliography{custom}
\bibliographystyle{acl_natbib}

\newpage
\appendix
\section*{Appendix}
\section{Motivation and Challenges}\label{apx:challenge}
\subsection{Motivation}
Besides the apparent benefits of a successful application of text classification for fault diagnosis, future benefits lie elsewhere. A software update can cause an entire fleet of vehicles to face the same problem in hours, which means that early fault diagnosis can allow efficient management of such cases. Furthermore, information sharing between workshops and markets can also be optimised. For example, some problems might occur earlier in e.g. Europe than Africa, which means that early training of technicians could provide substantial cost savings.

\subsection{Challenges}
Noteworthy is the fact the learning from other domains is not trivial. For example, what differentiates medical fault diagnosis is that it most often uses visual input, such as radiographs, making the state of the art not applicable to the automotive domain (and multi-linguality less related). The medical reports, on the other hand, comprise the diagnosis and are rather the objective~\cite{pavlopoulos2022diagnostic} instead of the input, as in our study. 
Also, using higher levels of the hierarchy would not help the expert, because they wouldn’t convey information regarding the needed parts or repair time. With that said, however, the hierarchical classification may be helpful for modelling purposes.

\section{Evaluation}\label{apx:eval}
Table~\ref{apx:catacc} presents the top-k categorical Accuracy of XLM-R and DistilBERT, which considers a prediction as true if the correct class is within the top-k predictions. Experimenting with k equal to three and five, XLM-R is better in both.

\begin{table}[h]
\caption{Top-3 and Top-5 categorical Accuracy of XLM-R and DistilBERT.}\label{apx:catacc} 
\begin{tabular}{lcc}\hline
 & DistilBERT & XLM-R \\\hline
Top-3 Accuracy & 75.4\% & \bf 77.9\% \\
Top-5 Accuracy & 79.2\% & \bf 82.6\% \\
\end{tabular}
\end{table}

Tables~\ref{apx:xlmr} and~\ref{apx:dbert} show the Precision, Recall, and F1 scores of XLM-R and DistilBERT, macro-averaged per zone. XLM-R has a better F1 score than DistilBERT for top and mid classes, due to its better Recall. However, in low classes DistilBERT is superior in all metrics, lifting also the overall performance (1st column of Table~\ref{apx:dbert}).

\begin{table}[ht]
\caption{Precision, Recall and F1 of XLM-R macro-averaged per zone. In bold the best results compared to the ones of DistilBERT in Table~\ref{apx:dbert}.}\label{apx:xlmr} \begin{tabular}{lcccc}\hline
 & \textbf{Total} & \textbf{Top} & \textbf{Mid} & \textbf{Low}  \\ \hline
Precision & 28.4\% & 74.5\% & 57.9\% & 20.4\% \\
Recall & 26.4\% & \textbf{80.2\%} & \textbf{63.0\%} & 16.6\% \\
F1 & 25.4\% & \textbf{77.0\%} & \textbf{59.3\%} & 16.2\% \\
\end{tabular}%
\end{table}
\begin{table}[h]
\caption{Precision, Recall and F1 of DistilBERT macro-averaged per zone. In bold the best results compared to ones of XLM-R in Table~\ref{apx:xlmr}.} \label{apx:dbert} \begin{tabular}{lcccc}\hline
& \textbf{Total} & \textbf{Top} & \textbf{Mid} & \textbf{Low}\\\hline 
Precision & \textbf{44.0\%} &\textbf{74.6}\% & \textbf{60.0\%} & \textbf{39.7\%} \\
Recall & \textbf{35.6\%} & 79.4\% & 60.1\% & \textbf{29.0\%} \\
F1 & \textbf{36.4\%} & 76.5\% & 58.5\% & \textbf{30.5\%} \\
\end{tabular}%
\end{table}%

\subsection{The unknown language}\label{apx:unknown}
When we use the original texts for samples whose translation is not available, DistilBERT gives 61.4\% in total and 37.4\%, 63.4\% and 81.2\% for low, mid and top classes respectively, which means that the low class is improved while mid and top are harmed.

\section{Confusion matrices}\label{apx:conf}
Tables~\ref{tab:cmat_distil} and~\ref{tab:cmat_xlmr} below show the confusion matrices for the DistilBERT model and the XLM-R model for the ten most frequent classes. The high values on the diagonal of the matrices imply that both models predict the top 10 classes well. 
\begin{table*}[h]
\caption{Confusion matrix for the top-10 classes for DistilBERT. Rows correspond to the true class and columns to the predicted class.}  \label{tab:cmat_distil}
\begin{tabular}{cc|c|c|c|c|c|c|c|c|c|c|c}
  & \multicolumn{1}{c}{} & \multicolumn{1}{c}{\textbf{5}}  & \multicolumn{1}{c}{\textbf{16}}  & \multicolumn{1}{c}{\textbf{12}} & \multicolumn{1}{c}{\textbf{2}} & \multicolumn{1}{c}{\textbf{0}} & \multicolumn{1}{c}{\textbf{41}} & \multicolumn{1}{c}{\textbf{159}} & \multicolumn{1}{c}{\textbf{140}} & \multicolumn{1}{c}{\textbf{64}} & \multicolumn{1}{c}{\textbf{90}}  \\\cline{3-12}
            & \textbf{5} & 2154 & 88 & 0 & 2 & 1 & 2 & 0 & 1 & 0 & 1 \\\cline{3-12}
            & \textbf{16} & 113 & 1534 & 0 & 0 & 0 & 0 & 0 & 0 & 0 & 1 \\\cline{3-12}
            & \textbf{12} & 0 & 0 & 1374 & 53 & 0 & 3 & 0 & 0 & 0 & 0 \\\cline{3-12}
            & \textbf{2} & 0 & 0 & 41 & 1181 & 0 & 2 & 0 & 0 & 0 & 0 \\\cline{3-12}
            & \textbf{0} & 1 & 0 & 1 & 1 & 1126 & 1 & 0 & 0 & 0 & 0 \\\cline{3-12}
            & \textbf{41} & 2 & 0 & 0 & 2 & 0 & 1081 & 0 & 0 & 2 & 0 \\\cline{3-12}
            & \textbf{159} & 1 & 1 & 0 & 1 & 0 & 0 & 735 & 128 & 0 & 1 \\\cline{3-12}
            & \textbf{140} & 4 & 0 & 1 & 1 & 0 & 2 & 161 & 762 & 0 & 1  \\\cline{3-12}
            & \textbf{64} & 10 & 1 & 0 & 0 & 0 & 0 & 0 & 2 & 890 & 0 \\\cline{3-12}
            & \textbf{90} & 2 & 1 & 0 & 1 & 0 & 0 & 1 & 2 & 0 & 645 \\\cline{3-12}
\end{tabular}
\end{table*}

\begin{table*}[h]
 \caption{Confusion matrix for the top-10 classes for XLM-R. Rows correspond to the true class and columns to the predicted class.} \label{tab:cmat_xlmr}
  \begin{tabular}{cc|c|c|c|c|c|c|c|c|c|c|c}
  & \multicolumn{1}{c}{} & \multicolumn{1}{c}{\textbf{5}}  & \multicolumn{1}{c}{\textbf{16}}  & \multicolumn{1}{c}{\textbf{12}} & \multicolumn{1}{c}{\textbf{2}} & \multicolumn{1}{c}{\textbf{0}} & \multicolumn{1}{c}{\textbf{41}} & \multicolumn{1}{c}{\textbf{159}} & \multicolumn{1}{c}{\textbf{140}} & \multicolumn{1}{c}{\textbf{64}} & \multicolumn{1}{c}{\textbf{90}}  \\\cline{3-12}
            & \textbf{5} & 2305 & 67 & 0 & 0 & 1 & 2 & 0 & 2 & 2 & 0 \\\cline{3-12}
            & \textbf{16} & 159 & 1595 & 0 & 2 & 0 & 0 & 0 & 0 & 1 & 1 \\\cline{3-12}
            & \textbf{12} & 0 & 1 & 1487 & 42 & 2 & 2 & 0 & 0 & 0 & 0 \\\cline{3-12}
            & \textbf{2} & 0 & 0 & 94 & 1216 & 0 & 1 & 0 & 0 & 0 & 0 \\\cline{3-12}
            & \textbf{0} & 0 & 0 & 1 & 1 & 1134 & 0 & 0 & 0 & 0 & 0 \\\cline{3-12}
            & \textbf{41} & 2 & 0 & 0 & 2 & 0 & 1083 & 0 & 0 & 1 & 0 \\\cline{3-12}
            & \textbf{159} & 1 & 0 & 0 & 1 & 0 & 0 & 794 & 117 & 1 & 0 \\\cline{3-12}
            & \textbf{140} & 7 & 1 & 1 & 0 & 0 & 2 & 184 & 790 & 1 & 1  \\\cline{3-12}
            & \textbf{64} & 12 & 2 & 0 & 0 & 0 & 1 & 0 & 1 & 906 & 0 \\\cline{3-12}
            & \textbf{90} & 3 & 0 & 0 & 0 & 0 & 2 & 0 & 1 & 0 & 660 \\\cline{3-12}
\end{tabular}
\end{table*}

\end{document}